\documentclass[]{article}

\usepackage{arxiv}

\usepackage[utf8]{inputenc} % allow utf-8 input
\usepackage[T1]{fontenc}    % use 8-bit T1 fonts
\usepackage{hyperref}       % hyperlinks
\usepackage[dvipsnames]{xcolor}
\hypersetup{
	colorlinks   = true,
	citecolor    = JungleGreen,
	allcolors    = PineGreen
}
\usepackage{url}            % simple URL typesetting
\usepackage{booktabs}       % professional-quality tables
\usepackage{amsfonts}       % blackboard math symbols
\usepackage{nicefrac}       % compact symbols for 1/2, etc.
\usepackage{microtype}      % microtypography
\usepackage{graphicx, float}
\usepackage[round]{natbib}
\usepackage{multicol}

\usepackage[greek,russian,english]{babel} 
\usepackage{stmaryrd} % semantic symbols
\usepackage{pifont}  % for ding symbols
\usepackage{float}
\usepackage{graphicx}
\usepackage{amsmath}
\usepackage{enumitem}
\usepackage{hyperref}

\usepackage[round]{natbib}

\usepackage{blindtext,tikz}
\usetikzlibrary{calc}

\title{Quantification and object perception in Multimodal Large Language Models and human linguistic cognition}

\author{ Raquel Montero$^1$,
	Natalia Moskvina$^1$,
	Paolo Morosi$^1$, 
	Tamara Serrano$^1$, 
	Elena Pagliarini$^1$ and 
	Evelina Leivada$^{1,2}$
}

\date{1. Universitat Autònoma de Barcelona\\ 2. Institució Catalana de Recerca i Estudis Avançats (ICREA)}

\renewcommand{\shorttitle}{Quantification and object perception in MLLMs}

\begin{document}

\maketitle

\begin{abstract}
Quantification has been proven to be a particularly difficult linguistic phenomenon for (Multimodal) Large Language Models (MLLMs). 
However, given that quantification interfaces with the logic, pragmatic, and numerical domains, the exact reasons for the poor performance are still unclear.
This paper looks at three key features of human quantification shared cross-linguistically that have remained so far unexplored in the (M)LLM literature: the ordering of quantifiers into scales, the ranges of use and prototypicality, and the biases inherent in the human approximate number system.
The aim is to determine how these features are encoded in the models' architecture, how they may differ from humans, and whether the results are affected by the type of model (thinking vs. instruct) and the language under investigation.
Results show that although thinking models showed a high accuracy in the numerosity estimation task and in the organization of quantifiers into scales, there are still key differences between humans and MLLMs across all model types, particularly in terms of ranges of use and prototypicality values.
This work, thus, paves the way for addressing the nature of MLLMs as semantic and pragmatic agents, while the cross-linguistic lens can elucidate whether their abilities are robust and stable across different languages.

\end{abstract}

% keywords 

\section{Introduction}\label{sec1}
Quantifiers reflect fundamental aspects of human cognition, including how abstract relationships of quantity and proportion are represented and processed. 
These concepts are critical for reasoning and natural language understanding, linking linguistic meaning with numerical cognition and symbolic thought \citep{chierchia2000,odic2015}.
As such, the study of quantification and object perception can serve  as a stringent testbed for assessing the true depth of language understanding in (Multimodal) Large Language Models (M)LLMs, and whether or not they can model the complexity that humans effortlessly manage \citep{szymanik2016}. 
While evidence has pointed out some problems in quantification in several of these models, both in visual and text-based tasks \citep{qiu2023,enyan2024,testoni2024}, other work has shown substantial similarities in quantifier usage between humans and (M)LLMs \citep{wong2025, kryvosheieva2025}.
Given the intricate system underlying human quantification as well as the variety of tasks deployed in previous work, determining the exact features that are accurately represented in the models' internal architecture and those that are not remains challenging.

One of the most influential linguistic theories to formalize the complexity present in natural language quantifiers is the \textit{Generalized Quantifier Theory} (GQT), which postulates that quantifiers denote relations between sets \citep{barwise1981generalized}.\footnote{
	Another competing account is the parameterized determiner approach \citep{romero2015conservativity} which analyzes quantifiers as gradable adjectives, that is, as relations between degrees and individuals.
	Truth-conditional approaches like GQT and the parametrized determiner approach have been criticized in the literature for not capturing the presence of prototypes and the gradient usage of quantifiers around those prototypes, but recent work by \citet{van2021} shows that truth-conditional approaches combined with a probabilistic pragmatic model can capture these properties as well as Prototype Theory.
	For the purposes of this paper, GQT will be followed, as is usually considered the standard analysis, but nothing essential in our claims hinges on this decision.}
According to GQT, quantifiers are analyzed as two-place predicates taking two sets as arguments and returning a truth value \citep{partee2012mathematical}.
For example, the expression ``All Ps are Q'' would be true if and only if the set of Ps is a subset of the set of Qs \eqref{ex-all-set-theory}.

\begin{enumerate}
	\item \label{ex-all-set-theory}
	$\llbracket$All Ps are Q$\rrbracket$ = 1 iff $P\subseteq Q$
\end{enumerate}

While the truth values of some quantifiers (e.g., \textit{all, some,} etc.) are context-independent \eqref{ex-lexical-entries-context-independent}, for other quantifiers (e.g., \textit{many, few}, etc.)  different truth conditions obtain depending on the context \eqref{ex-lexial-entries-context-dependent}.
For the latter type of quantifier the exact interpretation depends on the value assigned to the threshold $n$, which can be affected by factors such as expected frequency and situational knowledge \citep{van2014,greer2014}.

\begin{enumerate}[resume]
	\item \label{ex-lexical-entries-context-independent}
	\begin{enumerate}[label= \alph*), left=0.25em]
		\item $\llbracket$ all $\rrbracket$ = $\lambda P_{<e,t>}. \lambda Q_{<e,t>}. P \subseteq Q$
		\item $\llbracket$ some $\rrbracket$ = $\lambda P_{<e,t>}. \lambda Q_{<e,t>}. P \cap Q \neq \emptyset$
	\end{enumerate}
	\item \label{ex-lexial-entries-context-dependent}
	\begin{enumerate}[label= \alph*), left=0.25em]
		\item $\llbracket$ many $\rrbracket^c$ = $\lambda P_{<e,t>}. \lambda Q_{<e,t>}. |P \cap Q| > n_c$, for some large $n_c$
		\item $\llbracket$ few $\rrbracket^c$ = $\lambda P_{<e,t>}. \lambda Q_{<e,t>}. |P \cap Q| < n_c$, for some small $n_c$
	\end{enumerate}
\end{enumerate}

Based on these definitions, quantifiers can be ordered into lexical scales \eqref{ex-lexical-scales} according to their entailments relationships.
Stronger elements in the scale will entail weaker elements (e.g., all people came $\Rightarrow$ some people came),  but not \textit{viceversa}, (e.g., some people came $\not \Rightarrow$ all people came).
The lexical competition among these elements in the scale determines quantifier usage and interpretation.

\begin{enumerate}[resume]
	\item \label{ex-lexical-scales} Lexical scale: $<some, many, all>$
\end{enumerate}

For example, due to the competition between the elements in the scale, the meaning of a quantifier can, in certain contexts, be (pragmatically) enriched beyond its set theoretic definition \citep{horn1972}.
The quantifier \textit{some} in sentence \eqref{ex-some-scalar}, for instance, which is usually interpreted semantically as \textit{some and possibly all} can also be interpreted as meaning \textit{some but not all} \citep{grodner2010some}.
This type of enrichment, known in the literature as scalar implicature (SI), is typically attributed either to pragmatic reasoning guided by Gricean maxims \citep{horn1972,grice1975} or to a covert grammatical mechanism involving a silent ``only'' operator \citep{fox2007,chierchia2013}.\footnote{
	Gricean and Neo-Gricean accounts \citep{horn2014} view scalar implicatures as arising for specific lexical items.
	In contrast, Relevance Theory \citep{sperber1986} rejects the notion of default inferences, proposing instead that implicatures are context-driven and emerge only when they contribute to the overall relevance of the utterance.
}

\begin{enumerate}[resume]
	\item \label{ex-some-scalar}
	Some circles are blue.
	\begin{enumerate}[label= \alph*), left=0.25em]
		\item Some and possibly all circles are blue. 
		\item Some but not all circles are blue. 
	\end{enumerate} 
\end{enumerate}

The gradient nature of quantifier usage and their prototypical values \citep{van2014} emerge as a consequence of the non-categorical application of this (pragmatic) enrichment, which some authors have formalized in terms of Bayesian inferences, see \citep{van2021}.

In addition to the semantic-pragmatic mechanism described above, quantifier usage and comprehension depend on humans numerosity estimation skills.
In the literature two independent numerical cognitive systems have been proposed: a precise number system for counting small values (up to four), and an approximate number system for larger quantities \citep{feigenson2004}. 
These two systems are subject to biases associated with sensory properties such as object spacing, density, color and size, among others \citep{krueger1972,pauw2013,bertamini2018}.
Given that the use of quantifiers relies on these cognitive systems, these biases permeate into the quantificational system, and have repercussions on how humans produce quantifiers.

Human quantification is, thus, a very intricate linguistic and cognitive phenomenon, and the subpar performance of the models could stem from many different sources, ranging from a failing to replicate the logical meaning of the quantifiers, their contextual thresholds, generic interpretations, their ordering into scales, humans numerosity estimation skills, etc.

Moreover, the extent to which the reported problems on quantification are model-specific has not been systematically tested.
In particular, no study has compared the quantifier usage across models with different numerical skills.
Given that human quantification is directly link to numerical cognition and symbolic thought, by studying the performance of models with different mathematical abilities, a better understanding of the link between the two domains can be obtained.

Lastly, all previous literature on the interpretation of quantification in (M)LLMs has focused on the English language, raising the question of whether the empirical picture from this language obtains global support.
Studying the performance across typological diverse languages as well as languages from the same linguistic family is a necessary step to further our understanding regarding whether MLLMS posses a centralized conceptual representation that is externalized in many languages (as multilingual humans do), or whether there are language-specific neurons/linguistic maps for each language (family) \citep{tang2024language,wang2025}.
Previous work has shown that syntactic information is encoded in specific circuits and model units \citep{ferrando2024}, and some of these (e.g., agreement phenomena) show partial overlap cross-linguistically in a sample of 57 languages \citep{kryvosheieva2025}.
The extent to which these generalizations also hold in the semantic domain has, so far, not been explored.

This work fills these gaps in the literature by exploring three key research questions (RQ).
\textbf{RQ1}) \textbf{Why do some MLLMs struggle to use quantifiers in a human-like fashion?}
In particular, the paper focuses on three key aspects of quantification that have been so far not explored in the MLLM literature:
(i) the mental organization of quantifiers into ordered scales \eqref{ex-lexical-scales} \citep{horn1972,pezzelle2018};
(ii) the ranges of use and prototypical values  \citep{van2014,van2021,ramotowska2024}; and
iii) the cognitive biases of numerosity estimation in a broad sense \citep{krueger1972,pauw2013,bertamini2018}.
\textbf{RQ2)} \textbf{Are there differences in terms of performance across models with different numerical skills?}
To answer this question, the performance of thinking --also known as reasoning models-- (o4-mini and GLM-4.6V-thinking) and instruct --also called general purpose models-- (GPT-4o, Gemma 3 and Molmo 2) models is compared.
While the former has been trained via Reinforcement Learning on verifiable problems to create long chains of thought, the latter has not undergone that additional training.
Evidence shows that this additional fine-tuning step increases the models' accuracy in mathematics exams and other numerical tasks \citep{OpenAI2024}; hence, we expect the training to have affected the use of quantification as well. 
\textbf{RQ3)} \textbf{Are there differences in performance across languages?} This study expands the empirical domain of investigation to typologically diverse languages (English, Greek and Russian) as well as languages belonging to the same language family (Spanish, Italian and Catalan), to study the role of linguistic distance and the existence of linguistic maps in the models.

Given that natural language quantification integrates knowledge from multiple domains and modalities, the study of these questions presents an excellent opportunity to better understand the differences and gaps between statistical pattern recognition and genuine semantic interpretation. 
The present work, thus, advances our understanding regarding the cognitive plausibility and reasoning capabilities of computational models and contributes to the ongoing debate regarding their status as semantic and pragmatic agents \citep[among many others]{mitchell2023debate,saba2023stochastic,huetal2023,jian2024llms,lyre2024understanding,dentella2025language}.
The following section presents in more detail the methodology used to answer these questions.

\section{Methodology}
\label{sec-Methodology}

Participants are presented with visual stimuli consisting of black squares and white circles, and are asked to choose the quantifier that best describes the image \eqref{eg-production-task}.\footnote{
	\citet{testoni2024}'s work had images consisting of animals and non-animals.
	To simplify the task as much as possible, circles and squares are used instead.
}
In contrast to previous studies \citep{testoni2024, wong2025}, participants are additionally asked to choose the approximate proportion of squares in the image \ref{ex-blank-proportion}.
This question is included to be able to discern whether the difference between humans and models lies in their numerosity estimation skills (mapping from \ref{ex-blank-proportion} to the real number of objects in the image), or in the internal mapping of quantifiers to reported proportions (mapping from \ref{ex-blank-quantifier} to \ref{ex-blank-proportion}).

\begin{enumerate}[resume]
	\item \textbf{Production Task} \label{eg-production-task}
	\begin{enumerate}[label= \alph*), left=0.25em]
		\item Visual Stimuli: Image with $N$ squares and $100 - N$ circles. \label{ex-image}
		\item $[$Most/Many/Some/A few$]$ objects are squares.\label{ex-blank-quantifier}
		\item Approximate proportion of squares:  $[$11\%-20\%, 21\%-30\%...$]$ \label{ex-blank-proportion}
	\end{enumerate}
\end{enumerate} 

\begin{figure}[H]
	\centering
	\includegraphics[width=145pt]{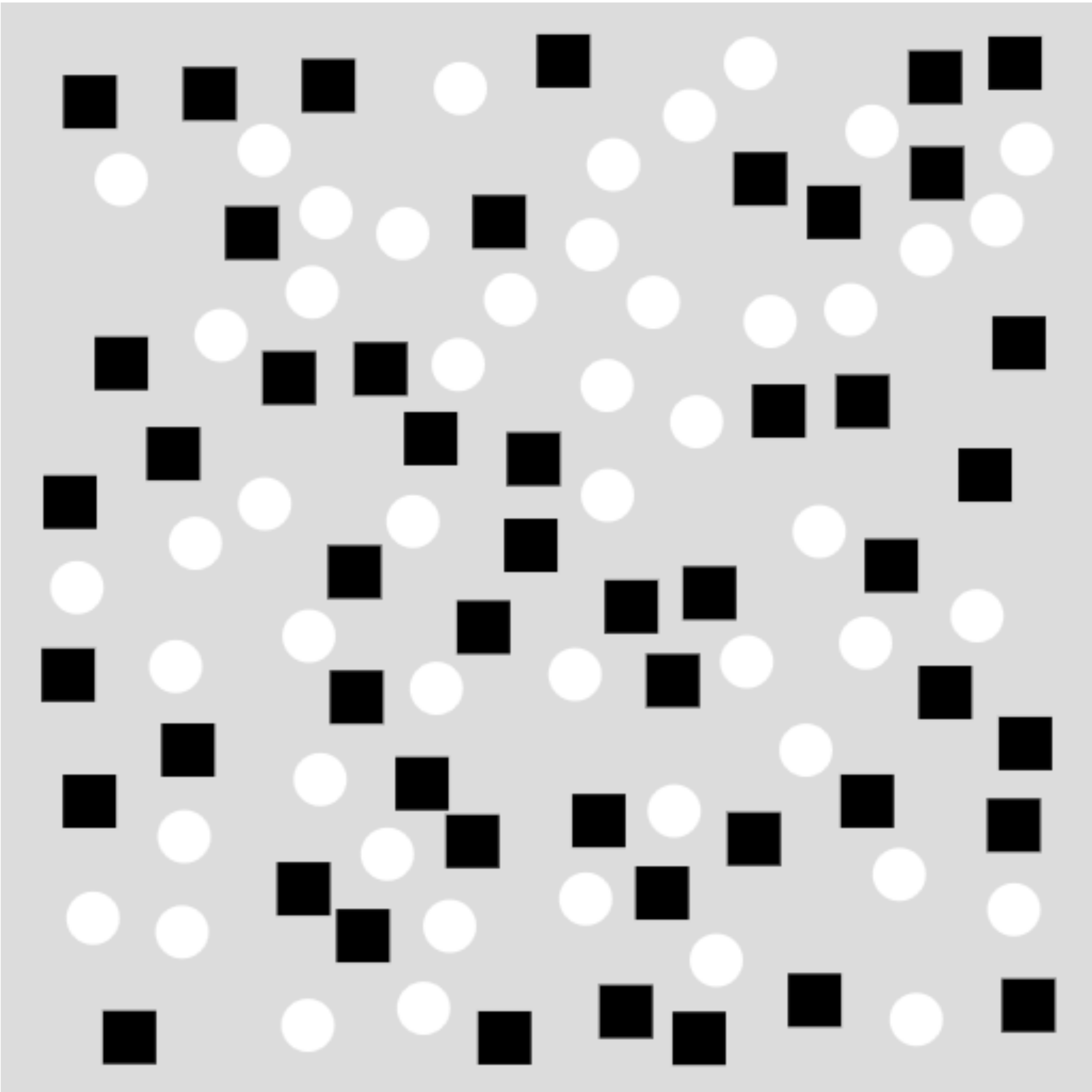}
	\caption{\label{fig-item} The experiment's visual stimuli.}
\end{figure}

In order to facilitate object identification by the models, the prompt includes information regarding the color and size of the objects in the image.
Moreover, in order to avoid positional biases \citep{pezeshkpour2024large} the lists of options as well as the images appear randomly ordered across trials. 
All images are created using JavaScript's p5 library, keeping the size of all the objects constant across items (25 pixels in width for the squares, and 25 pixels in diameter for the circles) and placing them (pseudo-)randomly, avoiding overlapping, in a 512x512 canvas.
Figure \ref{fig-item} shows an example. \footnote{The materials can be found at: \href{https://osf.io/uc2sv/overview?view\_only=86642ca4109741bca23c7219f4543ef8}{https://osf.io/uc2sv/overview?view\_only=86642ca4109741bca23c7219f4543ef8}}

The study is conducted in six languages: English, Greek, Russian, Spanish, Italian and Catalan.
The translations of the quantifiers are based on native speakers' intuitions who were instructed to match the meaning of the English quantifier as much as possible, both in terms of their semantic lexical entries and of the scalar implicatures they license.
The judgments are further confirmed through a thorough literature review of quantification in each language (\citet{giannakidou2012} for Greek, \citet{krasikova2011} for Russian, \citet{crisma2012} for Italian, \citet{marti2015} for Spanish, and \citet{brucart2002} for Catalan).
The final list of quantifiers is given in \eqref{ex-quantifiers}:

\begin{enumerate}[resume]
	\item \label{ex-quantifiers} \textbf{Quantifiers used in the study} 
	\begin{enumerate}[label= \alph*), left=0.25em]
		\item English: \textit{a few, some, many, most}
		\item Greek: \foreignlanguage{greek}{λίγα, κάποια, πολλά, τα περισσότερα}
		\item Russian: \foreignlanguage{russian}{несколько, некоторые, многие, большинство}
		\item Spanish: \textit{unos pocos, algunos, muchos, la mayoría}
		\item Italian: \textit{pochi, alcuni, molti, la maggior parte}
		\item Catalan: \textit{uns quants, alguns, molts, la majoria}
	\end{enumerate}
\end{enumerate}

The experiment was carried out in accordance with the Declaration of Helsinki and was approved by the Research Ethics Committee at UAB.
In total, 240 human participants were recruited via Prolific (40 from each language). 
Of these, 107 self-identify as female; 124 as male; 2 as non-binary; 1 as a trans man; and 6 did not specify. 
The ages of the participants ranged (in years) from 18 to 83.
The study was run online and was implemented in PCIbex \citep{zehr2018}.

Regarding the models, 5 different MLLMs are prompted via OpenAI's API \citep{opeAIAPI} and OpenRouter \citep{OpenRouter}:
GPT-4o \citep{gpto4}, o4-mini \citep{o4mini2025}, Gemma 3 (24B) \citep{gemma2025}, GLM-4.6V (106B) \citep{glm} and Molmo 2 (8B) \citep{molmo22026}.
Overall, the design aimed to have a balanced sample of instruct and thinking models, while at the same time having a representative sample of closed, open weights and fully open access models.
However, to date, there is no accessible fully open multimodal and multilingual reasoning model; hence the reason for only including an instruct model in the fully open-source category.
For closed models, the GPT family was chosen because it showed the highest performance in linguistic tasks in previous work \citep{yue2024large,dentella2025language}.
The final list of the models chosen and their characteristics can be found in Table \ref{table-models}.

\begin{table}[h]
	\centering
	\begin{tabular}{|c|c|c|}
		\hline
		Model Name 	& Type 	   & Source 		\\ \hline
		GPT-4o 		& Instruct & Closed			\\
		o4-mini 	& Thinking & Closed			\\
		Gemma 3 	& Instruct & Open-weights 	\\
		GLM-4.6V 	& Thinking & Open-weights 	\\
		Molmo 2 	& Instruct & Fully open 	\\ \hline
	\end{tabular}
	\caption{Models tested in the study.}\label{table-models}
\end{table}

Given that models have been shown to perform best at numerical tasks when the temperature is low, the temperature parameter (if available) was set to 0.
For the reasoning models, the reasoning effort parameter is left in medium.

In order to obtain a similar amount of data from humans and models, the study was run 40 times in each of the models. 
The final data analysis includes 4560 responses from humans, 4560 from Gemma 3, 4560 from GLM-4.6V, 4560 from o4-mini, 4211 from GPT-4o and 4533 from Molmo 2.
These last two models have less responses in the final data analysis, because they  sometimes responded \textit{I don't know} or something that did not comply with the task, and this type of response has been excluded.
A complete summary of the responses provided as well as the responses that were filtered out can be found in the online repository.

\section{Results and data analysis}
\label{sec-results}

\subsection{Approximate number system}
\label{subsec-results-counting}

\begin{figure}[h!]
	\centering
	\includegraphics[width=0.8\textwidth]{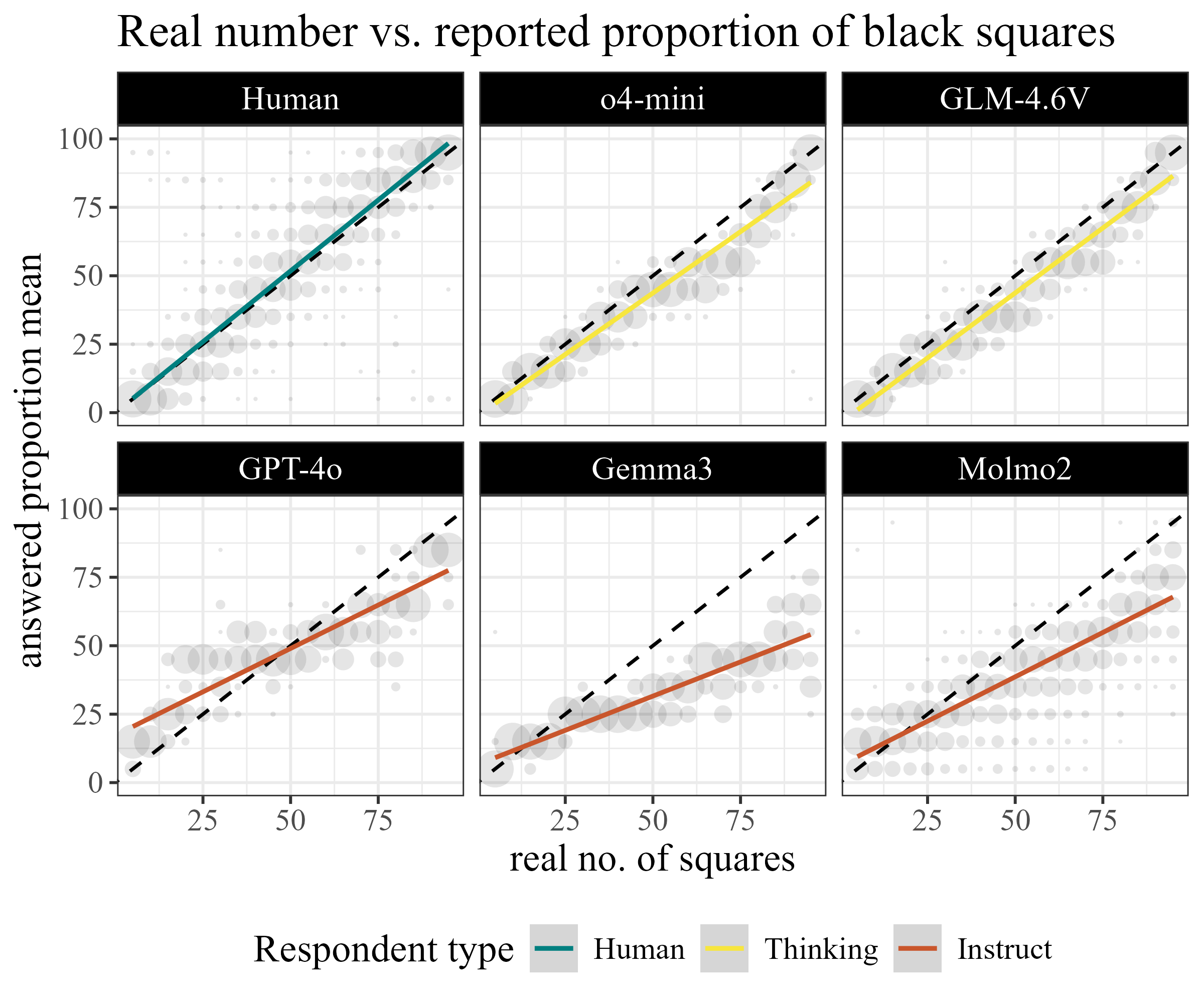}
	\caption{\label{fig-results-approximate-counting}Real vs. reported proportion of black squares in the stimuli. The size of the dot indicates the number of responses, the stripped line indicates the theoretically correct response, and the continuous lines, the model fit. Panels group responses based on the respondent. Thinking models are GLM-4.6V and o4-mini. Instruct models are GPT-4o, Gemma 3 and Molmo 2.}
\end{figure}

Figure \ref{fig-results-approximate-counting} shows the responses given by the participants regarding the perceived approximate proportion of black squares in the image ($y$-axis) in comparison to the real number of squares in that image ($x$-axis).

Generally, humans show a high accuracy when guessing the proportion of black squares in the images.
With respect to the models, thinking models (o4-mini and GLM-4.6V) outperformed general purpose models (GPT-4o, Gemma 3 and Molmo 2).
These later type tended to underestimate the proportion of black squares in the images when they were above 50.

In order to analyze these results, a linear-mixed effects regression model was run in R (version 4.5.1) using the packages lme4 \citep{lme4} and lmerTest \citep{lmerTest}.
The Reported Proportion was set as the dependent variable and the real Number of Squares and Type of Respondent (human/thinking/instruct) as independent variables.
Participants were added as crossed random effects.
The results showed a significant main effect of the Number of Squares ($df = 1$, $p <.001$) and Respondent Type ($df = 2$, $p < .001$) with instruct models choosing significantly higher proportions than humans ($\beta = 10, SE = 0.496,  t = 20.240 , p <.0001$) and thinking models choosing significantly lower proportions ($\beta = -2.457, SE = 0.525,  t= -4.680, p < .0001$).
Moreover, there was a significant interaction between the two effects  ($df = 2, p < 0.0001$).
In order to better understand it, a \textit{post-hoc} test is conducted using the \texttt{emtrends()} function with Tukey adjustment.
The results show that the slopes for all three types of respondents are differently from each other: humans vs. thinking ($\beta = 0.112, SE = 0.00535, df = 25541, t = 20.927, p <.0001$), human vs. instruct ($\beta = 0.442, SE = 0.442, df = 25543,  t = 87.304, p <.0001$) and thinking vs. instruct ($\beta = 0.330, SE = 0.00401, df = 25545, t = 82.294, p <.0001$).
However, thinking models performed closer to humans than instruct models, as can be seen when comparing their estimates: $\beta_{\text{human-thinking}}$ = 0.112 vs. $\beta_{\text{human-instruct}}$ = 0.330.

Overall, these results align with previous literature showing that reasoning models are better at numerical tasks than general purpose ones \citep{wei2022chain}.
Moreover, the results highlight that the reasons why some models deviate from humans in the way they use quantifiers when describing images may stem from differences in their approximate number systems.
The next section explores in more detail whether there are also differences in the meaning of quantification itself.

\subsection{Production of quantifiers}
\label{subsec-results-production}
\begin{figure}[h!]
	\centering
	\includegraphics[width=\textwidth]{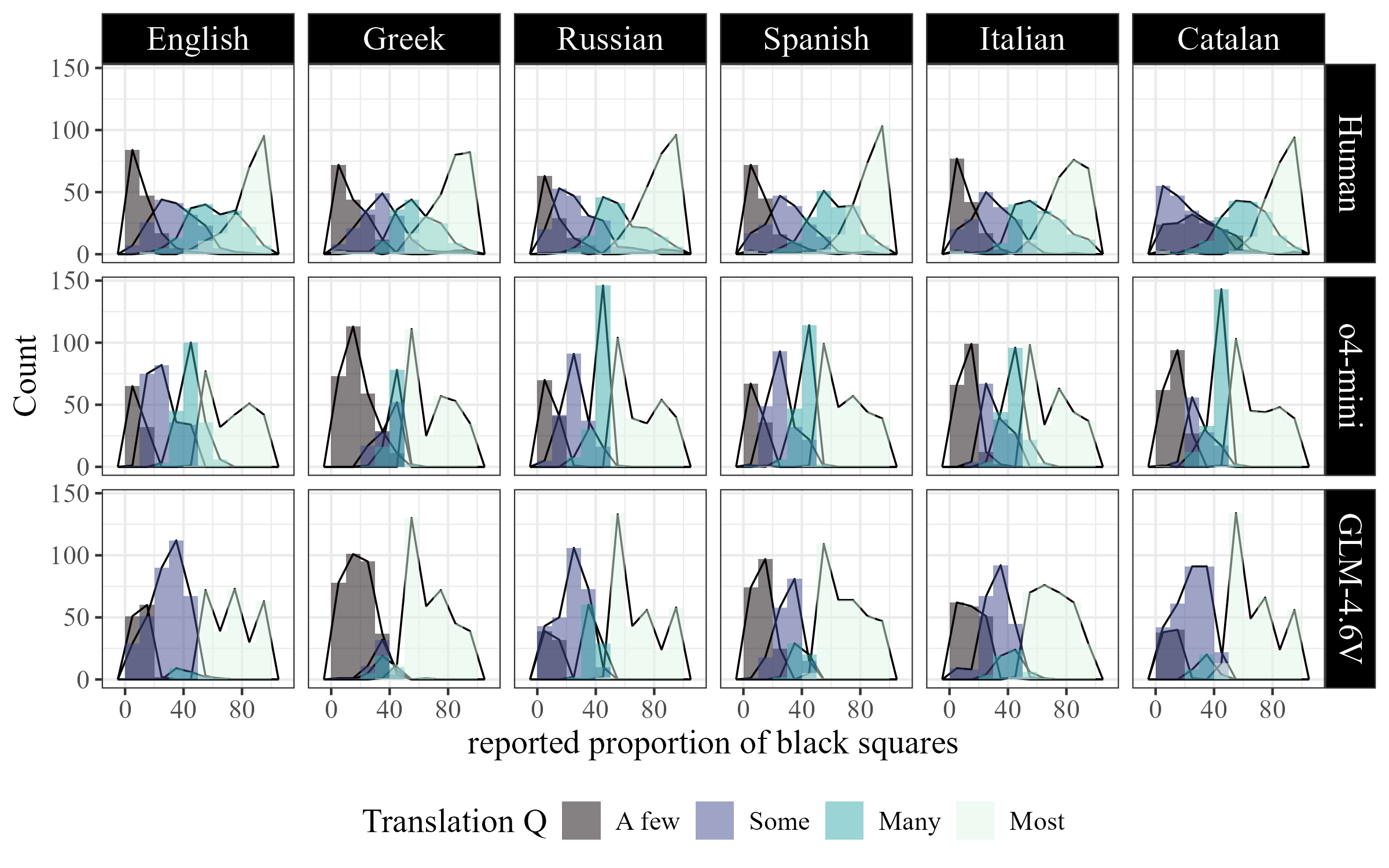}
	\caption{\label{fig-model-humans} Comparison of humans and thinking models in the production of quantifiers.  The first row shows the results from humans across the 6 different languages tested, the second row the results from o4-mini and the last row the results from GLM-4.6V.}
\end{figure}

This second subsection presents the results obtained for how humans and models mapped quantifiers into (perceived/reported) proportions.
The human benchmark can be found in Appendix \ref{appendix-cross-linguistic-benchmark}.
Figure \ref{fig-model-humans} shows the results of comparing humans' performance with that of thinking models.
The $x$-axis represents the reported proportion of black squares in the image and the $y$-axis the frequency with which the quantifier is selected for that chosen proportion.

Starting with the ordering of quantifiers into scales, the human data aligns with previous literature \citep{pezzelle2018, ramotowska2024}, and shows how across most of the languages tested, the quantifiers can be ordered following the scale $<\text{\textit{a few}, \textit{some}, \textit{many}, \textit{most}}>$. The quantifier \textit{a few} is used with lower magnitudes than the quantifier \textit{some}, \textit{some} is used with smaller proportions than \textit{many}, and \textit{many} with smaller ones than \textit{most}. 
Catalan seems to be an exception to this generalization, as the range of distributions of the quantifier \textit{uns quants} (`a few') and \textit{alguns} (`some') are the same.
One could formalize these by stating  that at the population level these two quantifiers are competing for the same position in the scale $<\text{\textit{a few}/\textit{some}, \textit{many}, \textit{most}}>$.\footnote{At the individual level there are differences in the ordering of the quantifiers: some individuals consistently use \textit{uns quants} as ``few'' and \textit{alguns} as ``some'' and for other speakers the opposite pattern is observed. Individual speakers may, hence, have different lexical representations.}

These orderings are, to a certain extent, also observed in thinking models, but there is also cross-linguistic and model specific variation in this respect.
For example, while o4-mini orders quantifiers according to the scale $<\text{\textit{a few}, \textit{some}, \textit{many}, \textit{most}}>$, GLM-4.6V follows  $<\text{\textit{a few}, \textit{some}/\textit{many}, \textit{most}}>$, the quantifier \textit{many} appearing in a subset of the contexts in which \textit{some} is used (i.e., $many$ is rarely chosen by the model, and the quantifier $some$ is the preferred option.).
Moreover, thinking models display cross-linguistic variation not observed in humans.
For example, GLM-4.6V chose predominantly the quantifiers \textit{a few} and \textit{most} in Greek and \textit{some} and \textit{most} in Catalan, a difference not present in the human data. 
Conversely, the existing human variation is not mirrored in the models. 
For instance, the variation observed in the human Catalan responses with respect to the quantifier \textit{a few} is not replicated in o4-mini.
Rather the quantifier \textit{uns quants} `a few' follows a similar distribution to the other languages.
These results, thus, show that thinking models, while ordering generally quantifiers in a distribution that resembles the human baseline, also show variation not present in humans and conversely, do not capture cross-linguistic differences observed in humans.

With regards to the ranges of use and prototypicality values, clear differences between humans and the two models can be observed. 
These differences become more evident when comparing the Mode --which can be used as a proxy of the most prototypical context for a given quantifier-- and the Interquantile Ranges (IQR).
For example, while the Mode for the quantifier \textit{most} is 91\%-100\% cross-linguistically in the human data, the Modes in o4-mini and GLM-4.6V range between $51\%-70\%$. 
Regarding the IQRs, the models' values tend to be either much larger or much narrower than those of humans.
For instance, for the quantifier \textit{many}, 50\% of the human data lies within the 40\%-80\% interval, with an IQR of 40; whereas the data from thinking models for this quantifier lies in the 30\%-50\% interval, with an IQR of 20.
These differences indicate problems in finding the quantifiers' contextual thresholds \textit{n} \eqref{ex-lexial-entries-context-dependent}; a  result in line with previous research \citep{enyan2024}.

\begin{figure}[h!]
	\centering
	\includegraphics[width=0.9\textwidth]{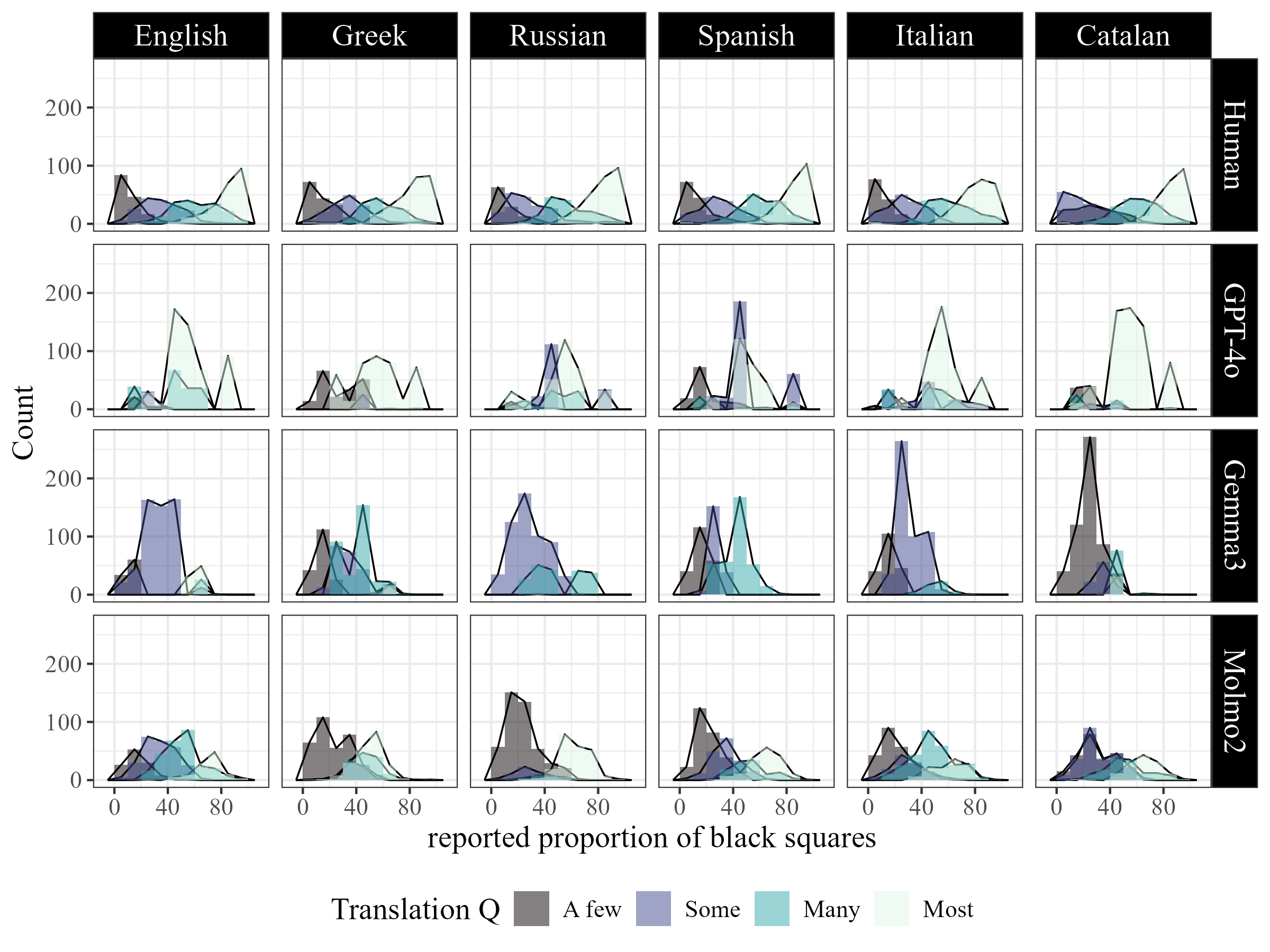}
	\caption{\label{fig-quantfifiers-general} Comparison of humans and instruct models in the production of quantifiers.}
\end{figure}

With respect to instruct models, the tested MLLMs also show differences not observed in the human baseline, but the patterns are different from the ones present in thinking models.
The results for instruct models can be found in Figure  \ref{fig-quantfifiers-general}.

Starting with the ordering of quantifiers into scales, in contrast to thinking models, instruct models do not follow an order that resembles the human baseline.
For example, in Gemma 3, 50\% of the responses for the quantifier \textit{most} in English are within the 50\%-70\% interval, and for the quantifier \textit{many} within the 60\%-70\% interval.
That is, the quantifier \textit{most} is not used with bigger proportions than \textit{many}.
Similarly, in GPT-4o, 50\% of the data for the quantifier \textit{many} in Spanish lies within the range 10\%-70\%, but within 40\%-50\% for the quantifier \textit{some}.
The only exception to these observations is Molmo 2 in English, were the quantifiers do follow the order $<$ \textit{a few}, \textit{some}, \textit{many}, \textit{most} $>$.

Moreover, when looking at the usage of the quantifiers, instruct models do not comply with the set theoretic definitions. 
For example, GPT-4o uses the quantifier  \textit{most} across almost all the available proportions,  and 50\% of the data lies in the interval 40\%-70\%.
However, according to the set theoretic definition \eqref{ex-set-theory-most} the quantifier should only be used with proportions bigger than 50\%.

\begin{enumerate}[resume]
	\item \label{ex-set-theory-most} $\llbracket$ most $\rrbracket$(P)(Q) = 1 iff $|P\cap Q| > |P-Q|$
\end{enumerate} 

Similarly, regarding the prototypicality values and ranges of use there are key differences between instruct models and humans.
For example, the quantifier \textit{most} has a Mode of $91\%-100\%$ in humans, but of $51\%-60\%$ in instruct models.
Furthermore, the distributions in instruct models are different from humans, insofar they have a narrower range of usage.
For example, the quantifier \textit{many} has an IQR of 30 in humans, but of 20 in the models. 

These results indicate thus, that although thinking models organize quantifiers in a manner that resembles the human pattern more, both types of models show differences with respect to prototypicality and ranges of use.

In order to test if the differences are statistically significant, six multinomial regressions (one for each language\footnote{Given that there are differences in the quantifiers across languages (Appendix \ref{appendix-cross-linguistic-benchmark}) strictly speaking the dependent variable (the quantifier chosen) is different in each language, hence the need to run separate analyses for each language.}) are run using the \textit{nnet} package \citep[v.7.3.20]{nnet}, with the Quantifier (\textit{a few, some, many, most}) as the dependent variable, and the Reported Proportion of Squares and the Type of Respondent (human, thinking, instruct) as independent variables, with humans and the quantifier \textit{most} set as the reference levels.
Model selection is performed by comparing the Akaike's Information Criterion (AIC) and residual deviance of each model (models with one main effect, models with two main effects, models with an interaction), and the one with the lowest scores are selected.
The models with the interaction obtained a lower AIC and deviance score, and were thus selected for further data analysis.
Across all of these models, there was a main effect of Type of Respondent ($p <.05$) and an interaction between the Type of Respondent and the Reported Proportion of Squares ($p <.05$).
Given that the independent variable, Respondent Type, has three categories, in order to better understand the nature of the interaction, a \textit{post-hoc} test is conducted using the \texttt{emtrends()} function with Tukey adjustment.
The results of the relevant conditions can be found in Table \ref{tb-p-values}.

\begin{table}[h]
	\centering
	\begin{tabular}{c c c c c}
		% HUMANS vs Thinking
		\multicolumn{5}{c}{\textbf{Humans vs. Thinking}}\\\hline
		& Most 			& Many    		  	  & Some  				& A few\\\hline
		English  & 0.2293    		& \textbf{0.0057}     &  \textbf{$<$0.0001}	& \textbf{0.0163} \\
		Greek    & 0.8823  		    & \textbf{0.0304}  	  & 0.8973 				& \textbf{0.0005}\\
		Russian  & \textbf{0.0004}  & \textbf{0.0001}  	  & \textbf{$<$0.0001}  &\textbf{$<$0.0001} \\
		Spanish  & 0.5487   		& \textbf{0.0423}     & \textbf{$<$0.0001}  & \textbf{0.0001}\\
		Italian  & 0.2477   		& \textbf{0.0204}     &	\textbf{0.0214}		& \textbf{0.0221}\\
		Catalan  & \textbf{0.0053}  & 0.9886  			  & \textbf{$<$0.0001}	& \textbf{$<$0.0001} \\[3pt]
		% HUMANS vs instruct
		\multicolumn{5}{c}{\vspace{0.05 in}\textbf{Humans vs. Instruct}}\\\hline
		& Most 			   & Many    		    & Some  			 & A few\\\hline
		English  & \textbf{$<$0.0001}  & \textbf{0.0002}   	& \textbf{0.3981}	 & 0.1155 \\
		Greek    & \textbf{$<$0.0001}  &  \textbf{$<$0.0001}& \textbf{0.0252}    & 0.0764\\
		Russian  & \textbf{0.0019}     & \textbf{0.002}     & 0.0675		 	 & 0.9064\\
		Spanish  & \textbf{0.003}      & \textbf{$<$0.0001} & \textbf{0.003}	 & 0.7633\\
		Italian  & \textbf{0.0020}     & \textbf{$<$0.0001}	& \textbf{0.0238}	 & 0.99997\\
		Catalan  & \textbf{$<$0.0001}  & \textbf{$<$0.0001} & \textbf{0.0013}	 & \textbf{$<$0.0001}\\\hline
		
	\end{tabular}
	\caption{\label{tb-p-values}$p$-values of the \textit{post-hoc} test comparing humans with thinking and instruct models. Cells in bold indicate the difference is statistically significant.}
\end{table}

As can be seen, in the case of instruct models, the low magnitude quantifier \textit{a few} shows a more human-like behavior than the other quantifiers.
In the case of thinking models, the opposite pattern is observed. Thinking models performed closer to the human base-line in the case of the quantifier \textit{most}. 

Lastly, it is worth highlighting that the quantifiers \textit{many} and \textit{some} are statistically different across most model types and languages, indicating a clear difficulty when encoding the meaning of these quantifiers (we will return to this point in Section \ref{sec-discussion}).

These results, thus, show that the performance of the models is significantly different from humans in all the models tested, and that there is an effect of language. 
However, English was not the best performing language. 
This result aligns with other current work in the literature showing how English, contrary to expectation, is not the best performing language cross-linguistically \citep{weissweiler2023counting,kim2025one}, see Section \ref{sec-discussion} for further discussion on these points.

Overall, our results suggest that the models tested map quantifiers to proportions in a non-human-like fashion.
%In the next subsection, the embeddings of the models are further explored to corroborate these observations.

\section{Discussion}
\label{sec-discussion}

This paper has investigated three research questions: \textbf{RQ1)} Why do MLLMs struggle with quantification? 
\textbf{RQ2)} Are there differences in terms of performance across models with different levels of numerical accuracy (e.g., reasoning vs. non-reasoning)?
\textbf{RQ3)} Do the same patterns of behavior observed for English also emerge cross-linguistically?

Regarding RQ1, it has been shown that the tested MLLMs have problems not only with object identification and numerosity estimation, but also with the concept of quantification itself.
In particular, the experimental results show that the models' use of the quantifiers do not always comply with set theoretic definitions, and do not align with humans in terms of the ranges of use, prototyicality values and the ordering of quantifiers into scales.
We hypothesize that these problems may more generally stem from the distributional semantic approach upon which the models are built on.
LLMs rely on statistical distributional patterns found in natural language to build their representations.
While such an approach has been very successful in capturing the semantics of lexical words, it encounters difficulties when trying to capture the meaning of topic independent words, such as negation or numerals \citep{abrusan2018}.
Indeed, it has been shown that (M)LLMs face challenges with the interpretation of negation \citep{nadeem2024} and arithmetic \citep{guo2025your}. 
Given that quantification builds on notions such numerosity estimation and cardinality differences, these problems may have permeated into the representation of quantification.
In text-based tasks, the systems might be able to replicate superficial correlations, but in multimodal tasks where the meaning of the quantifiers is essential to carry the specific tasks, these problems might be more salient, as demonstrated in this paper.

Regarding RQ2, the paper has shown that there are clear differences across models types.
By comparing thinking with instruct models, it has been shown that the ordering of quantifiers into scales and their set theoretic definitions can be captured by the thinking model, in stark contrast to instruct ones.
As briefly commented in the introduction, the key difference between these model types is that the former is further trained via Reinforcement Learning on verifiable problems to create long chains of thought \citep{OpenAI2024}.
This type of training has been shown to improve the numerical accuracy of the models in mathematical and coding tasks; a result that is also replicated in this study in the vision domain, as o4-mini significantly improved object identification and counting in comparison to GPT-4o (Section \ref{subsec-results-counting}).
Given that o4-mini excels at numerical tasks, it is not surprising that in the domain of quantifiers (which interface with the numerical domain), these models outperformed general purpose ones.
However, RL seems to be insufficient to fully capture the meaning of the quantifiers, thinking models still behave differently from humans, mostly in terms of typicality values and ranges of use.
We hypothesize that this might be due to the methodology relying on a ground truth answer, but lacking training on the intra-speaker variability that emerges when (scalar) implicatures are used by humans.

Lastly, regarding RQ3, the paper has expanded the empirical domain of investigation to a wide range of languages with different writing systems (Cyrillic, Greek and Latin-based scripts) and belonging to different language families (Balto-Slavic, Hellenic, Romance and Germanic).
It has been shown that: i) the models do not display the exact same behavior across the tested languages; ii) although there are differences across languages, the models do not replicate human behavior in any of languages studied in the paper; and iii) the quantifiers \textit{many} and \textit{some} are cross-linguistically the hardest quantifiers for the models.
The results, thus, suggest that the models have different internal representations for the tested quantifiers across different languages (rather than a centralized system), and that language distance might play only a marginal role in how quantification is encoded in the models' architecture. 
Attesting to this, the Romance languages tested in the study did not show a common pattern of behavior.
Moreover, although model accuracy differed from language to language, it was interesting that the quantifier \textit{many} was the hardest one throughout the tested languages.
This result was unexpected given that this quantifier relies on the same linguistic mechanisms as the other quantifiers, and in set theoretic terms is not more complex than quantifiers such as \textit{most} which also requires access to the cardinalities of the sets.
The pattern is even more puzzling given that the highest performing quantifier across languages in instruct models is \textit{a few}; a compositional variant of \textit{few}, which some authors have hypothesized could be analyzed as meaning ``not many''\eqref{ex-meaning-few} \citep{solt2006}.

\begin{enumerate}[resume]
	\item \label{ex-meaning-few} $\llbracket$ few $\rrbracket$ = $\neg$ $\llbracket$ many $\rrbracket$ 
\end{enumerate}

Moreover, when considering language acquisition, it has been shown that \textit{some} is cross-linguistically mastered earlier than \textit{most} \citep{katsos2016}; this pattern has often been attributed to the relative semantic simplicity of \textit{some}.
However, in our results, no significant specific facility for the quantifiers \textit{some} was observed.

These results, thus, highlight important differences in how quantifier are mapped into proportions across different model types, and show-case patterns of behavior that are unexpected if the models' were encoding human-like semantic knowledge, as no current linguistic theory would predict such patterns.

\section{Conclusion}
\label{sec-conclusion}

This paper has studied the semantic and pragmatic abilities of MLLMs in the realm of quantification. 
In line with previous research \citep{collacciani2024,testoni2024, enyan2024}, we found that the models use quantifiers in a manner that differs from human baselines.
Addressing the so far unanswered why question (i.e. why do models differ from humans?), our experiments have delved into the exact source of these problems by comparing how different aspects of quantification are encoded and utilized by the models.
Moreover, the paper has extended the domain of investigation to other languages and model types; furthering our understanding of how quantification is encoded in these models more broadly.

These results, thus, shed light on which specific aspects of quantification might be difficult for these models to replicate, and pave the way for capturing the nature and the possible limitations of MLLMs as semantic and pragmatic agents.

\newpage

\bibliographystyle{apalike}
\bibliography{references}

\newpage

\appendix

\section{Appendix: Cross-linguistic benchmark }
\label{appendix-cross-linguistic-benchmark}

\begin{figure}[h!]
	\includegraphics[width=\textwidth]{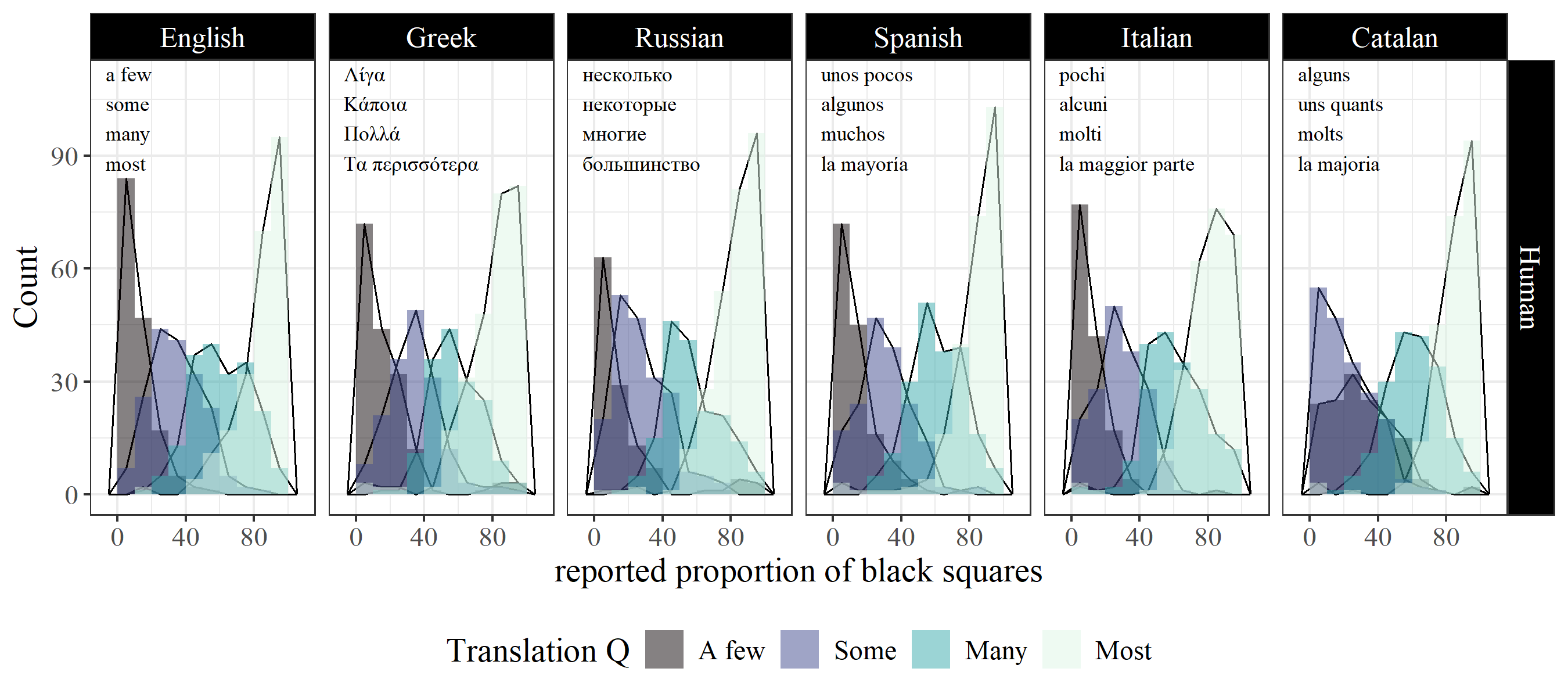}
	\caption{\label{fig-cross-linguistic}Cross-linguistic benchmark.}
\end{figure}

Figure \ref{fig-cross-linguistic} shows the results obtained for humans.
On the $x$-axis appears the reported proportion of black squares in a given image, and on the $y$-axis the count with which each quantifier is chosen.

Across all languages, with the exception of Catalan, the quantifiers are ordered following the lexical scale: $<$\textit{a few}, \textit{some}, \textit{many}, \textit{most}$>$. \textit{A few} represents a lower magnitude than \textit{some}, \textit{some} represents a smaller magnitude that \textit{many}, and \textit{many} represents a smaller magnitude that \textit{most}.

The main difference comes in the exact shape the distribution each quantifier has, both in terms of its range of usage and its typicality values.
For example, for the quantifier \textit{some} the Mode is 21-30\% in English, Spanish and Italian but 11-20\% in Russian, 31-40\% in Greek and 0-10\% in Catalan.
For the quantifier \textit{a few} the Interquartile Range (IQR) is 20 in English and Greek, 10 in Russian, Spanish and Italian, and 30 in Catalan.

In order to test if the differences are statistically significant, for each of the quantifiers a logistic regression model (glm) is run in R (v.4.5.1),  with Quantifier (Q$_n$/Other) as the dependent variable and Reported Proportion of black squares and Language -- with reference level set to English-- as fixed factors, if they improve model fitting.
Results show that the quantifier \textit{most} is statistically different from the English equivalent in Greek ($\beta= 0.55, SE= 0.17, z=3.3, p<0.001$), Russian ($\beta=0.48, SE= 0.17, z=2.8, p=0.005$) and Italian ($\beta=0.41, SE = 0.16, z=2.5, p=0.01$).
For the quantifier \textit{many} there is no effect of language, and including it as a fixed factor does not improve the model fit ($p = 0.1976$)--as estimated by comparing the logLikelihoods of the models using the \texttt{anova()} function--.
Finally, the quantifier \textit{a few} is significantly different from English across all the languages: Greek ($\beta = -0.94851, SE =  0.38176,  z =-2.485, p = 0.013$), Russian ($\beta = -2.16572,  SE =  0.36194, z =  -5.984, p<0.0001$), Spanish ($\beta = -1.02015, SE = 0.38604, z = -2.643, p = 0.008$), Italian ($ \beta = -0.99831, SE = 0.38991, z =  -2.560, p=0.011$) and Catalan ($\beta=-2.81468, SE= 0.34897 , z= -8.066, p<0.0001$).
This difference reflects that in English this quantifier is used more frequently with smaller values (mean$_{\text{English}}$ = 12) than in the other languages (e.g., mean$_{\text{Russian/Greek}}$ = 17, mean$_{\text{Spanish}}$ = 14).

\end{document}